\pdfoutput=1

\documentclass[11pt]{article}

\usepackage[final]{acl}

\usepackage{times}
\usepackage{latexsym}

\usepackage[T1]{fontenc}

\usepackage[utf8]{inputenc}

\usepackage{microtype}

\usepackage{inconsolata}

\usepackage{graphicx}

\usepackage{booktabs}
\usepackage{url}
\usepackage{graphicx}
\usepackage{booktabs}
\usepackage{xcolor}
\usepackage{comment}
\usepackage{amsmath}
\usepackage{amsfonts}
\usepackage{upgreek}
\usepackage{siunitx}
\usepackage{multirow}
\usepackage{makecell}

\usepackage{natbib}
\usepackage{xspace}
\usepackage{soul}
\usepackage{adjustbox}
\usepackage{wrapfig}
\usepackage{arydshln}
\usepackage[normalem]{ulem}
\usepackage{tabularx} 
\usepackage{csquotes}

\usepackage{array}
\newcolumntype{L}[1]{>{\raggedright\let\newline\\\arraybackslash\hspace{0pt}}m{#1}}
\newcolumntype{C}[1]{>{\centering\let\newline\\\arraybackslash\hspace{0pt}}m{#1}}
\newcolumntype{R}[1]{>{\raggedleft\let\newline\\\arraybackslash\hspace{0pt}}m{#1}}
\newcolumntype{Y}{>{\centering\arraybackslash}X}

\title{Traffic Light or Light Traffic? Investigating Phrasal Semantics in Large Language Models}

\author{
  \textbf{Rui Meng}$^1$, \textbf{Ye Liu}$^1$, \textbf{Lifu Tu}$^1$, 
  \\ \textbf{Daqing He}$^2$, \textbf{Yingbo Zhou}$^1$, \textbf{Semih Yavuz}$^1$ \\
$^1$ Salesforce Research, $^2$ University of Pittsburgh\\
  { \texttt{ruimeng@salesforce.com}} 
}

\begin{document}
\maketitle
\begin{abstract}
Phrases are fundamental linguistic units through which humans convey semantics. This study critically examines the capacity of API-based large language models (LLMs) to comprehend phrase semantics, utilizing three human-annotated datasets. 
We assess the performance of LLMs in executing phrase semantic reasoning tasks guided by natural language instructions and explore the impact of common prompting techniques, including few-shot demonstrations and Chain-of-Thought reasoning.
Our findings reveal that LLMs greatly outperform traditional embedding methods across the datasets; however, they do not show a significant advantage over fine-tuned methods. The effectiveness of advanced prompting strategies shows variability. We conduct detailed error analyses to interpret the limitations faced by LLMs in comprehending phrase semantics\footnote{Related code and data can be found at \url{https://github.com/memray/llm_phrase_semantics/}}.

\end{abstract}

\section{Introduction}
Understanding phrase semantics presents unique challenges for Artificial Intelligence due to the compositionality and ambiguity inherent in natural languages. The same set of words can yield phrases with vastly different meanings, exemplified by phrase pairs such as ``traffic light'' versus ``light traffic'' and ``hard drive'' versus ``drive hard''. Understanding phrase semantics~\cite{2013semevalphrase,datasetturney,datasetbird,datasetppdb} involves reasoning the structure and meaning of the combined words and aligning these symbol with real-world concepts commonly understood by humans. Common tasks used to assess a model's ability to comprehend phrase semantics include identifying similar phrases or estimating the semantic similarity between phrase pairs.

Traditional methods for representing phrase semantics generally rely on either embedding-based strategies or fine-tuning approaches. Embedding-based methods~\cite{phrasebert,spanbert,sentencebert,simcse,fctmodel} generate vector representations for phrases, using vector similarity as a proxy for semantic relatedness. However, this approach often oversimplifies the complex relationships between phrases. On the other hand, fine-tuning methods~\cite{phrasebert,PiC} adapt a pretrained model (like BERT) to a specific task, such as classification or regression, but require extensive task-specific data.

Large language models (LLMs) excel by generalizing to new tasks with minimal instructions, an advantage in data-scarce situations. Innovations such as in-context few-shot learning~\cite{dong2022survey,cahyawijaya2024llms} and Chain-of-Thought prompting~\cite{wei2022chain,nguyen2023cof,ling2024deductive} have shown promise in transferring knowledge to novel tasks and enhancing reasoning skills. The potential of LLMs in phrase semantic understanding remains underexplored and warrants investigation.

This study evaluates the performance of LLMs in phrase semantics tasks by focusing on their instruction-following abilities. Our contributions are threefold:
(1) We analyze LLMs' processing of phrase semantics, particularly their alignment with human linguistic understanding -- a crucial yet underexplored aspect;
(2) We benchmark LLMs against a range of traditional methods to determine if LLMs can effectively perform semantic reasoning tasks solely based on instruction-following and various prompting techniques; 
(3) We provide a detailed error analysis to elucidate the reasons behind the failures of current LLMs in specific scenarios.

\begin{table*}[ht!]
\renewcommand\arraystretch{0.8}
\setlength\tabcolsep{3pt} 
\begin{center}
    \scriptsize
    \begin{tabular}{C{1cm}L{6cm}C{1cm}L{7cm}}
        \toprule
        \textbf{Task} & \makecell[c]{\textbf{Prompt}} & \makecell*[{{p{1cm}}}]{\textbf{Ground-truth}} & \makecell[c]{\textbf{Output}} \\ \midrule
        \makecell*[{{C{1cm}}}]{Turney\\+CoT} 
        & \makecell*[{{p{6cm}}}]{Given 5 word candidates (delimited by $\mid$), you are tasked to answer which word is most semantically similar to a given phrase. \\
        ~~~~Please respond with the word only, without any explanation. \\
        ~~~~Among 5 words "proverb $\mid$ sabre $\mid$ stockpiling $\mid$ moonlighter $\mid$ jigsaw", which word is most similar to the phrase "saber saw"? Let's think step by step, briefly explain the meaning of each word/phrase, and end the response with a new line that only contains the answer.} 
        & \makecell*[{{C{1cm}}}]{jigsaw} 
        & \makecell*[{{p{7cm}}}]{
        - Sabre: A type of sword with a curved blade.\\
        - Stockpiling: Accumulating a large amount of goods or materials.\\
        - Jigsaw: A tool used for cutting precise curves in wood, metal, or other materials.\\
        - Moonlighter: A person who has a second job in addition to their main one, typically done at night.\\
        - Proverb: A short, well-known saying that expresses a general truth or piece of advice.\\
        ~~~~Jigsaw} \\ \hline
        
        \makecell[c]{BiRD\\+CoT} & \makecell*[{{p{6cm}}}]{Given two phrases, you are tasked to predict the semantic relatedness of them. \\
        ~~~~Now, given two "access road" and "road access", what is their semantic relatedness? Please only output a number within the range [0, 100]. \\
        ~~~~Let's think step by step, briefly explain the meaning of each phrase, and end the response with a new line that only contains the estimated number of semantic relatedness.}
        & \makecell*[{{C{1cm}}}]{0.684}
        & \makecell*[{{p{7cm}}}]{``Access road'' typically refers to a road that allows people to reach a main road or destination.\\
        ``Road access'' generally means the ability to reach or use a road. It implies that there is a way to get onto the road from a property, area, or another road.\\
        Given the close relationship between the physical entity of an access road and the concept of having road access, the relatedness is high.\\
        ~~~~95.} \\ \hline
        

        \makecell[c]{PiC\\+CoT} & \makecell*[{{p{6cm}}}]{Given a pair of phrases sharing the same context, you are tasked to answer whether the two phrases have the same semantic meaning. Now here is the test: \\
        ~~~~- On the local circuit, Faiz emerged champion in 3 events and obtained medals in both the Foil and Sabre U-12 events.\\
        ~~~~- On the community competitions, Faiz emerged champion in 3 events and obtained medals in both the Foil and Sabre U-12 events.\\
        ~~~~Please answer whether the phrase pair "local circuit" and "community competitions" has the same semantic meaning. Let's think step by step, briefly explain the similarity/difference between the two phrases, and end the response with only Yes or No).} 
        & \makecell*[{{C{1cm}}}]{yes}
        & \makecell*[{{p{7cm}}}]{
        ~~~~The phrase ``local circuit'' typically refers to a series of competitions that are held within a specific local area. It implies a sequence of events that participants travel around to compete in.\\
        ~~~~On the other hand, ``community competitions'' suggests contests or events that are organized within a community, which could be a town, city, or a smaller locality. This term emphasizes the community aspect and may not necessarily imply a series of events or the same structured nature as a circuit.\\
        ~~~~While both phrases indicate events that are local and not of a larger national or international scale, ``local circuit'' suggests a more organized series of events, whereas ``community competitions'' could be less structured and not part of a circuit.\\
        ~~~~No. } \\
        \bottomrule
    \end{tabular}
    \caption{Examples of instructions with CoT and the corresponding \texttt{GPT-4-Turbo} outputs on three tasks.}
    \vspace{-2.5em}
    \label{table:phrase_semantics_examples}
\end{center}
\end{table*}

\section{Methods}
We compare three types of models to evaluate their performance in understanding phrase semantics:

\noindent\textbf{Embedding similarity based methods}: these models encode phrases into dense vectors and measure semantic relatedness through vector similarity. For classification, the phrase with the highest vector similarity is selected, while for regression tasks, the similarity score itself is used as the prediction.

\noindent\textbf{Fine-tuning based methods}: these approaches involve adding a prediction head to a pretrained language model, fine-tuning it to capture the nuances of specific semantic tasks. In this study, results are reported for a classification dataset, PiC-PS, where a training split is available.

\noindent\textbf{Natural language instruction based methods}: Leveraging instruction-tuned models like GPT-3~\cite{llm_few} and FLAN~\cite{flan}, these methods guide language models by providing task instructions as part of the input. Techniques such as in-context few-shot demonstration~\cite{gpt3} and chain-of-thought prompting (CoT)~\cite{cot,emergent} are employed to enhance models' task-specific reasoning capabilities. When CoT prompting is enabled, we guide the model to execute semantic reasoning, i.e. explain the meaning of each phrase and their differences before making predictions, which is different from math or logical reasoning tasks~\cite{gsm8k,bigbench} that care about the soundness of the reasoning procedure.

\section{Experiment Setting}

To evaluate phrase semantics understanding, three datasets were utilized:

\noindent\textbf{Turney}~\cite{datasetturney}: Comprising 2,180 examples derived from WordNet, this dataset pairs one bigram phrase (query phrase) with five unigrams, where one unigram is a synonym of the bigram, and the others are related distractors. As a multiple-choice classification task, it challenges the model to identify the synonym from the unigrams, with performance measured by accuracy.

\noindent\textbf{BiRD}~\cite{datasetbird}: This is a bigram relatedness dataset using the Best–Worst Scaling (BWS) annotation. In contrast to common annotation techniques that use a discrete 0 to 5 scale, BWS is expected to provide more reliable and discriminating annotations. Specifically, it employs comparative annotations, where annotators are given multiple items at a time and asked to select which item is the best and worst. BiRD consists of 3,345 pairs of bigram phrases, each including a pair of phrases and a human rating of similarity between 0 and 1. The Pearson correlation coefficient between the model predictions and human judgments is reported.

\noindent\textbf{PiC-PS} (Phrase-in-Context - Phrase Similarity)~\cite{PiC}: PiC-PS is a binary classification task and it contains 2,000 examples for testing. The model is tasked to predict whether two noun phrases are semantically similar or not in the same context. The challenge of this task is that, without taking the contextual information into account, very likely the two phrases are interpreted as synonyms. The accuracy score is reported.

For Turney and BiRD, we replicate baselines following \cite{phrasebert}. For PiC-PS, we follow \cite{PiC} and reuse the reported scores. We select two API-based LLMs: \texttt{GPT-3.5-Turbo} (\textit{gpt-3.5-turbo-1106}) and \texttt{GPT-4-Turbo} (\textit{gpt-4-1106-preview}). Sample data and prompts are detailed in Table~\ref{table:phrase_semantics_examples} and \ref{table:apendix-example-turney-bird}. If the OpenAI API fails to produce valid responses (e.g., system refusal messages), we retry up to 10 times or assign default values (-1 for BiRD and \textit{INVALID} for others).
The potential of few-shot learning to enhance LLM performance was also considered. Given the absence of a development set for Turney and BiRD, 10 examples from each test split were allocated as a makeshift development set, leaving 2,170 and 3,335 examples for final testing, respectively.

\section{Results and Analyses}

\subsection{Main Results}
\begin{table}[ht!]
    \footnotesize
    \centering
    \renewcommand\arraystretch{0.9}
    \begin{tabular}{lcccc}
    \toprule
        \textbf{Model} & \multicolumn{2}{c}{\makecell{\textbf{Turney}\\\textbf{(Accuracy)}}} & \multicolumn{2}{l}{\makecell{\textbf{BiRD}\\\textbf{(PearsonCC)}}} \\ \midrule
        \multicolumn{2}{l}{\textbf{Baselines} (Embedding)} & \\ \midrule
        BERT & 42.5 && 0.445 \\ 
        BERT-Large & 42.5 && 0.476 \\ 
        SpanBERT & 20.8 && 0.257 \\ 
        SpanBERT-Large & 41.0 && 0.330 \\ 
        Phrase-BERT & \underline{57.1} && \underline{0.689} \\ \midrule
        \multicolumn{2}{l}{\textbf{LLMs} (Instruction)} & +CoT & & +CoT \\ \midrule
        \texttt{GPT-3.5T}, 0-shot  & 87.0 & 83.4
        & 0.719 & 0.689 \\ 
        \texttt{GPT-3.5T}, 2-shot  & 87.0 & 76.0
        & 0.637 & 0.639 \\ 
        \texttt{GPT-3.5T}, 4-shot  & \underline{\textbf{87.6}} & 78.6
        & 0.636 & 0.632 \\ 
        \hdashline
        \texttt{GPT-4T}, 0-shot  & 86.1 & 86.2
        & 0.750 & 0.724 \\ 
        \texttt{GPT-4T}, 2-shot  & 86.8 & 86.4
        & 0.759 & 0.733 \\ 
        \texttt{GPT-4T}, 4-shot  & 87.5 & \underline{86.6}
        & \underline{\textbf{0.761}} & \underline{0.738} \\ 
        \bottomrule
    \end{tabular}
    \caption{Model performance on Turney and BiRD. Best score in each group/task is \underline{underlined}/\textbf{boldfaced}.}
    \label{table:turney_bird}
\end{table}

The results of phrase semantic similarity are presented in Table~\ref{table:turney_bird} and \ref{table:pic}. The evaluation of phrase semantic similarity across the datasets shows that instruction-based methods employing LLMs set new benchmarks (Turney from 57.1 to 87.6, BiRD from 0.689 to 0.761, PiC-PS from 69.3 to 73.5). However, the highest score on each dataset comes with different models or configurations. 

On the Turney benchmark, the previous best accuracy score (57.1) was achieved by an embedding-based method using Phrase-BERT and cosine similarity. Both LLMs under the basic prompting significantly outperform Phrase-BERT (+30 points), primarily due to enhanced language understanding capabilities. Furthermore, more improvements can be attained (GPT-3.5T +0.6, GPT-4T +1.4, no CoT) by providing few-shot examples within the context, allowing the model to glean more task-specific information for refining its predictions.

\begin{table}[!ht]
    \centering
    \renewcommand\arraystretch{0.9}
    \footnotesize
    \begin{tabularx}{0.5\textwidth}{Xcc}
    \toprule
        \textbf{Model} & \multicolumn{2}{c}{\textbf{PiC-PS Accuracy}} \\ \midrule
        \textbf{Baselines} & \textbf{M1: Embedding} & \textbf{M2: Fine-tuned} \\ \midrule
        BERT & 64.1 & 68.9 \\ 
        SpanBERT & 64.0 & 66.9 \\ 
        SpanBERT-Large & \underline{66.3} & \underline{69.3} \\ 
        SentenceBERT & 60.3 & 62.6 \\ 
        PhraseBERT & 63.4 & 66.1 \\ 
        SimCSE & 62.5 & 66.7 \\ \midrule
        \textbf{LLMs} & \textbf{M3: w/o CoT} & \textbf{M4: w/ CoT} \\ \midrule
        \texttt{GPT-3.5T}, 0-shot & 62.1 & 66.3  \\ 
        \texttt{GPT-3.5T}, 2-shot & 70.8 & 62.0  \\ 
        \texttt{GPT-3.5T}, 4-shot & 69.9 & 63.4  \\ 
        \texttt{GPT-3.5T}, 8-shot & 67.6 & 63.3  \\ 
        \hdashline
        \texttt{GPT-4T}, 0-shot & \underline{\textbf{73.7}} & 64.4  \\ 
        \texttt{GPT-4T}, 2-shot & 72.8 & 70.7  \\
        \texttt{GPT-4T}, 4-shot & 73.5 & \underline{72.3}  \\
        \texttt{GPT-4T}, 8-shot & 72.7 & 72.0  \\  \bottomrule
    \end{tabularx}
    \caption{Model performance on PiC-PS.}
    \vspace{-1em}
    \label{table:pic}

\end{table}

BiRD directly evaluates the alignment between human annotators and models on phrase similarities, which is measured by Pearson correlation coefficient. It may favor embedding-based methods since they were trained by optimizing the similarities between phrases~\cite{phrasebert}. LLMs output similarities in natural language, similar to the way that humans annotate the data~\cite{datasetbird}. The result shows that LLMs still outrun baselines by a clear margin. Similar to Turney results, few-shot examples show beneficial impacts on GPT-4T but CoT hurts both models. One possible reason is that the longer prompts potentially cause ``distractions'' to the models. For example, we observe many more bad-format cases for GPT-3.5T when CoT prompting is enabled. 


PiC-PS is a context-dependent test. The task is arguably more challenging as the difference between them can be subtle and models need to comprehend the context for support. The previous best score (69.3) is achieved by a fine-tuned SpanBERT-Large with 7k training examples. The best LLM setting only attains a minor improvement (73.7). The few-shot demonstration benefits \texttt{GPT-3.5-Turbo} the most when two examples are used, but it worsens the performance of \texttt{GPT-4-Turbo}. Additionally, the result shows that the CoT prompting leads to degradation in general. The models appear to give reasonable interpretations and very few invalid responses are detected.

\subsection{Error Analyses}
\label{sec:error-analyses}
LLMs demonstrate state-of-the-art performance across three phrase semantics benchmarks without any fine-tuning. However, advanced prompting techniques yield inconsistent results. Our goal here is to conduct thorough analyses to discern the reasons behind these failures.

\noindent\textbf{Failure of CoT}: 
We have observed that semantic CoT prompting generally impacts performance negatively. A benefit of CoT prompting is its ability to make the model’s reasoning processes transparent, allowing for detailed examinations of its predictions. In this study, we analyze 50 Turney incorrect predictions made by \texttt{GPT-4-Turbo}, using 4-shot and CoT to investigate the behaviors of LLMs.

Our analysis reveals three primary categories of errors:
\textbf{1. Erroneous selection despite correct analysis (27/50)}: The model correctly explains individual phrases -- like ``inkpot'' as a small container for ink and ``vessel'' as a container for holding liquids—but fails to select the most relevant phrase. For instance, despite explaining both terms accurately, the model erroneously selects ``vessel'' as most similar to ``ink bottle''.
\textbf{2. Unknown concepts (13/50)}: For example, it does not understand the phrase ``magic eye'' as referring to a photoelectric cell or a tuning indicator in radios. Thus the model mistakenly associates it with ``thaumaturgy'' (magic spells) instead of ``photocell''.
\textbf{3. Label ambiguity (7/50)}: Phrases with multiple meanings pose significant challenges. ``Small beer'' can denote either a minor issue or a type of beer. The model struggles to differentiate whether ``brew'' or ``trivia'' is more appropriate, given both could semantically align with ``small beer''. Additional error types account for the remaining cases (3/50).

\begin{figure}[h]
\vspace{-0.5em}
\begin{center}
\includegraphics[width=0.49\textwidth]{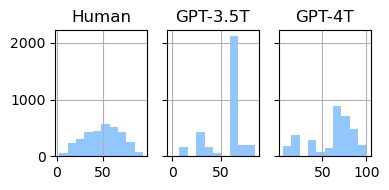}
\end{center}
\vspace{-1em}
\caption{Histograms of BiRD phrase-pair similarities assigned by human annotators, GPT-3.5-turbo and GPT-4-turbo (4-shot demonstration, no CoT prompting).}
\vspace{-1em}
\label{fig:bird-distribution}
\end{figure}

\noindent\textbf{Alignment to human semantic preference}: We analyze the distribution of phrase-pair similarities as assessed by human annotators and two LLMs, depicted in Figure~\ref{fig:bird-distribution}. The similarities assigned by human annotators typically follow a normal distribution. \texttt{GPT-3.5-Turbo}, on the other hand, frequently assigns a similarity value of 65.0 -- observed 1,725 times -- which may suggest a training bias. This atypical tendency is also noted under other test conditions. \texttt{GPT-4-Turbo} demonstrates a broader range of similarity assignments, yet both models tend to assign higher similarity scores than those given by human annotators. The distributions observed in CoT variants are similarly skewed. These findings indicate that LLMs are not yet fully aligned with human preferences regarding phrase similarity assessments.

\noindent\textbf{Failure to follow instructions}: In our tests, there were relatively few instances (less than 10 per test) where failures were due to invalid output formats. However, when Chain-of-Thought (CoT) prompting was enabled, the performance of GPT-3.5-Turbo on the Turney dataset deteriorated significantly. Specifically, the incidence of invalid outputs increased dramatically from 0/5/7 to 825/1,354/1,224 under the 0/2/4-shot settings. Moreover, in 37/159/136 of these cases, the model merely repeated the query phrases. The model often produced a sentence (e.g., ``the word most similar to the phrase `elephant bird' is: aepyornis'') instead of the expected single candidate phrase, necessitating post-processing (results shown in Table~\ref{table:turney_bird}). Despite changes in prompt settings, GPT-4-Turbo consistently generated valid outputs.


\section{Conclusion}
We investigated the performance of LLMs on phrase semantics to assess their alignment with human preferences. Despite strong performance across benchmarks, our findings reveal significant opportunities for improvement. Key areas include ensuring consistent adherence to instructions, crafting better prompts to enhance semantic reasoning, and integrating external knowledge to resolve ambiguities in phrase interpretation. This study inspects the alignment of LLMs with human understanding through the lens of phrase semantics, shedding light on the specific challenges and limitations these models face. It also inspires further exploration into diverse perspectives to assess the cognitive differences between AI agents and humans.

\newpage

\section{Limitations}

\paragraph{Model Dependency:} The research only utilizes two API-based large language models: \texttt{GPT-3.5-Turbo} and \texttt{GPT-4-Turbo}. They represent a specific subset of available LLMs, and their findings may not generalize to other models or newer versions.

\paragraph{Dataset and Evaluation Constraints:} Our experiments were conducted using three specific datasets, which may not encompass the full spectrum of phrase semantics diversity. The inherent ambiguity in phrase semantics can render benchmarking both challenging and debatable. Additionally, the BiRD dataset relies on human ratings of phrase pairs, introducing a degree of subjectivity and potential biases in the annotations. 

\paragraph{Prompt Dependence:} The study heavily relies on the effectiveness of prompting techniques, including few-shot demonstrations and CoT prompting. Models may be sensitive to the exact wording and structure of prompts, which can lead to variability in model performance. 

\bibliography{custom}

\appendix
\newpage

\section{Appendix}
\label{sec:appendix}
Examples show the complete collection of prompts used in this study and the corresponding model responses from GPT-4-Turbo.

\begin{table*}[!ht]
\renewcommand\arraystretch{1.0}
\begin{center}
    \scriptsize
    \begin{tabular}{C{1cm}lC{0.1cm}l}
        \toprule
        \textbf{Task} & \makecell[c]{\textbf{Prompt}} & \textbf{Ground-truth} & \makecell[c]{\textbf{Output}} \\ \midrule
        Turney & \makecell*[{{p{6cm}}}]{Given 5 word candidates (delimited by $\mid$), you are tasked to answer which word is most semantically similar to a given phrase. \\
        ~~~~Please respond with the word only, without any explanation. \\
        ~~~~Among 5 words "device $\mid$ keypad $\mid$ nihilism $\mid$ vestal $\mid$ machine", which word is most similar to the phrase "computer keyboard"?} & keypad & keypad \\ \hline

        \makecell[c]{Turney \\+ 2-shot} & \makecell*[{{p{6cm}}}]{Given 5 word candidates (delimited by $\mid$), you are tasked to answer which word is most semantically similar to a given phrase. \\
        ~~~~Please respond with the word only, without any explanation. \\
        ~~~~For example: \\
        ~~~~- Given phrase "street person" and words "thoroughfare $\mid$ individual $\mid$ dosser $\mid$ sectarianism $\mid$ paraquat", the answer is: dosser\\
        ~~~~- Given phrase "good story" and words "advantage $\mid$ sportsmanship $\mid$ cloudburst $\mid$ funny $\mid$ narrative", the answer is: funny \\
        ~~~~Among 5 words "device $\mid$ keypad $\mid$ nihilism $\mid$ vestal $\mid$ machine", which word is most similar to the phrase "computer keyboard"?} & keypad & keypad \\ \hline

        \makecell[l]{Turney \\+ 2-shot \\+ CoT} & \makecell*[{{p{6cm}}}]{Given 5 word candidates (delimited by $\mid$), you are tasked to answer which word is most semantically similar to a given phrase. \\
        ~~~~For example: \\
        ~~~~- Given phrase "street person" and words "thoroughfare $\mid$ individual $\mid$ dosser $\mid$ sectarianism $\mid$ paraquat", the answer is: dosser\\
        ~~~~- Given phrase "good story" and words "advantage $\mid$ sportsmanship $\mid$ cloudburst $\mid$ funny $\mid$ narrative", the answer is: funny \\
        ~~~~Among 5 words "device $\mid$ keypad $\mid$ nihilism $\mid$ vestal $\mid$ machine", which word is most similar to the phrase "computer keyboard"?\\
        ~~~~Let's think step by step, briefly explain the meaning of each word/phrase, and end the response with a new line which only contains the answer.} & keypad & \makecell*[{{p{6cm}}}]{~~~~- Vestal: Referring to or characteristic of a vestal virgin; chaste, pure, or related to the Roman goddess Vesta.\\~~~~- Machine: An apparatus using mechanical power and having several parts, each with a definite function and together performing a particular task.\\~~~~- Nihilism: The rejection of all religious and moral principles, often in the belief that life is meaningless.\\- Device: A thing made or adapted for a particular purpose, especially a piece of mechanical or electronic equipment.\\~~~~- Keypad: A set of buttons arranged in a block or "pad" which bear digits, symbols, or alphabetical letters and are used for entering data into a computer system or other device.\\ ~~~~keypad } \\ 
        \hline

        BiRD & \makecell*[{{p{6cm}}}]{Given two phrases, you are tasked to predict the semantic relatedness of them. Note that you are granted to have the capability to predict semantic relatedness between two specific terms. \\
        ~~~~Now, given two "access service" and "accessible location", what is their semantic relatedness? \\
        ~~~~Please only output a number within the range [0, 100].} & 0.465 & 70 \\ \hline

        \makecell[c]{BiRD \\+ 2-shot} & \makecell*[{{p{6cm}}}]{Given two phrases, you are tasked to predict the semantic relatedness of them. Note that you are granted to have the capability to predict semantic relatedness between two specific terms. \\
        ~~~~For example: \\
        ~~~~- The semantic relatedness of "world class" and "class world" is: 42.0\\
        ~~~~- The semantic relatedness of "world class" and "premier" is: 36.4 \\
        ~~~~Now, given two "access service" and "accessible location", what is their semantic relatedness? \\
        ~~~~Please only output a number within the range [0, 100].} & 0.465 & 58.7 \\ \hline

        \makecell[l]{BiRD \\+ 2-shot \\ + CoT} & \makecell*[{{p{6cm}}}]{Given two phrases, you are tasked to predict the semantic relatedness of them. Note that you are granted to have the capability to predict semantic relatedness between two specific terms. \\
        ~~~~For example: \\
        ~~~~- The semantic relatedness of "world class" and "class world" is: 42.0\\
        ~~~~- The semantic relatedness of "world class" and "premier" is: 36.4 \\
        ~~~~Now, given two "access service" and "accessible location", what is their semantic relatedness? \\
        ~~~~Please only output a number within the range [0, 100]. \\
        ~~~~Let's think step by step, briefly explain the meaning of each phrase, and end the response with a new line which only contains the estimated number of the semantic relatedness. It does not have to be a precise numerical score. Simply let me know a rough estimate based on your understanding.} & 0.465 & \makecell*[{{p{6cm}}}]{~~~~The phrase "access service" typically refers to the ability to use or obtain a service, which could be anything from internet access to customer support. It implies the provision of a service that can be used or reached.\\ ~~~~On the other hand, "accessible location" generally means a place that can be easily reached or entered, often used in the context of being user-friendly for individuals with disabilities or simply easy to get to for the general public.\\ ~~~~While both phrases involve the concept of accessibility, they apply to different contexts one to services, and the other to physical locations. However, there is a thematic overlap in the idea of being able to reach or use a service or location.\\ ~~~~Considering this thematic overlap, but also the difference in application (service vs. location), I would estimate the semantic relatedness of the two phrases to be moderate. Here's the estimated number for their semantic relatedness:\\ ~~~~55.0} \\ 
        \bottomrule
    \end{tabular}
    \caption{Prompts used in this study for Turney and BiRD and the corresponding outputs from \texttt{GPT-4-Turbo}.}
    \label{table:apendix-example-turney-bird}
\end{center}
\end{table*}

\begin{table*}[!ht]
\renewcommand\arraystretch{1.0}
\begin{center}
    \scriptsize
    \begin{tabular}{clC{1cm}l}
        \toprule
        \textbf{Task} & \makecell[c]{\textbf{Prompt}} & \textbf{Ground-truth} & \makecell[c]{\textbf{Output}} \\ \midrule
        
        PiC & \makecell*[{{p{7cm}}}]{Given a pair of phrases sharing the same context, you are tasked to answer whether the two phrases have the same semantic meaning.
        Now here is the test, in the two sentences below: \\
        ~~~~- However, James Alfred was not convicted in this or in a prior case of a similar nature.\\
        ~~~~- However, James Alfred was not convicted in this or in a preceding game of a similar nature.\\
        ~~~~The phrase pair "prior case" and "preceding game" has the same semantic meaning is (only output Yes or No):} & no & no \\ \hline

        \makecell[c]{PiC \\+ 2-shot} & \makecell*[{{p{7cm}}}]{Given a pair of phrases sharing the same context, you are tasked to answer whether the two phrases have the same semantic meaning.
        Now here is the test, in the two sentences below: \\
        ~~~~Here are 2 examples: \\
        ~~~~Example\#1, in the two sentences below: \\ 
        ~~~~- Carty attended a Saturday morning acting class at Phildene Stage School from the age of four, which had an attached agency.\\
        ~~~~- Carty attended a Saturday morning acting class at Phildene Stage School from the age of four, which had an accompanying business that represents actors.\\
        ~~~~phrases describe a business connected to the Stage School, with the primary function of representing actors.The phrase pair "attached agency" and "accompanying business that represents actors" has the same semantic meaning is: Yes\\
        ~~~~Example\#2, in the two sentences below: \\
        ~~~~- after theo's apparent death, she decides to leave first colony and ends up traveling with the apostles.\\
        ~~~~- after theo's apparent death, she decides to leave original settlement and ends up traveling with the apostles.\\
        ~~~~Both phrases refer to an initial establishment or habitation, but "first colony" carries specific connotations of being part of a series of colonies or a larger colonization effort, whereas "original settlement" is a more neutral term for any first habitation.The phrase pair "first colony" and "original settlement" has the same semantic meaning is: No\\
        Now here is the test, in the two sentences below: \\
        ~~~~- However, James Alfred was not convicted in this or in a prior case of a similar nature.\\
        ~~~~- However, James Alfred was not convicted in this or in a preceding game of a similar nature.\\
        ~~~~Please answer whether the phrase pair "prior case" and "preceding game" has the same semantic meaning?\\
        ~~~~Please only respond with Yes or No.} & no & no \\ \hline

        \makecell[l]{PiC \\+ 2-shot \\ + CoT} & \makecell*[{{p{7cm}}}]{Given a pair of phrases sharing the same context, you are tasked to answer whether the two phrases have the same semantic meaning. Now here is the test, in the two sentences below: \\
        ~~~~Here are 2 examples: \\
      ~~~~Example\#1, in the two sentences below: \\ 
      ~~~~- Carty attended a Saturday morning acting class at Phildene Stage School from the age of four, which had an attached agency.\\
      ~~~~- Carty attended a Saturday morning acting class at Phildene Stage School from the age of four, which had an accompanying business that represents actors.\\
      ~~~~Whether the phrase pair "attached agency" and "accompanying business that represents actors" has the same semantic meaning? \\
      Let's think step by step: Both phrases describe a business connected to the Stage School, with the primary function of representing actors.\\
      So the answer is: Yes\\
      ~~~~Example\#2, in the two sentences below: \\
      ~~~~- after theo's apparent death, she decides to leave first colony and ends up traveling with the apostles.\\
      ~~~~- after theo's apparent death, she decides to leave original settlement and ends up traveling with the apostles.\\
      ~~~~Whether the phrase pair "first colony" and "original settlement" has the same semantic meaning?\\
      ~~~~Let's think step by step: Both phrases refer to an initial establishment or habitation, but "first colony\
      " carries specific connotations of being part of a series of colonies or a larger colonization effort, whereas "original settlement" is a more neutral term for any first habitation.\\
      So the answer is: No\\
      Now here is the test, in the two sentences below: \\
      ~~~~- However, James Alfred was not convicted in this or in a prior case of a similar nature.\\
      ~~~~- However, James Alfred was not convicted in this or in a preceding game of a similar nature.\\
      ~~~~Please answer whether the phrase pair "prior case" and "preceding game" has the same semantic meaning. \\
      ~~~~Let's think step by step, briefly explain the similarity/difference between the two phrases, and end the response with only Yes or No).} & no & \makecell*[{{p{5cm}}}]{~~~~The phrase "prior case" refers to a legal matter or situation that has occurred before the current one, suggesting a context of legal proceedings or court cases. On the other hand, "preceding game" implies a sports or recreational activity that took place before another event. The key difference lies in the context implied by "case", which is legal, and "game", which is recreational or competitive. Therefore, the two phrases do not have the same semantic meaning.\\ ~~~~No.} \\
        \bottomrule
    \end{tabular}
    \caption{Prompts used for calling LLMs on PiC-PS and the corresponding outputs from \texttt{GPT-4-Turbo}.}
    \label{table:apendix-example-PiC}
\end{center}
\end{table*}

\end{document}